\documentclass{article} 
\usepackage{nips13submit_e,times}
\usepackage{hyperref}
\usepackage{url}
\usepackage{graphicx} 
\usepackage{subcaption} 
\usepackage{tabularx}

\usepackage{algorithm}
\usepackage{algorithmic}

\usepackage{times, amsmath, epsfig, graphicx,url, bm, multirow}
\usepackage{amssymb}
\usepackage{epstopdf}

\usepackage{tikz}
\usetikzlibrary{bayesnet}

\DeclareMathOperator*{\argmax}{arg\,max}

\title{A Generative Product-of-Filters Model of Audio}

\author{
Dawen Liang\thanks{ This work was performed while Dawen Liang was an intern at Adobe Research.} \\
Columbia University\\
New York, NY 10027\\
\texttt{dliang@ee.columbia.edu} \\
\And
Matthew D. Hoffman \\
Adobe Research \\
San Francisco, CA 94103\\
\texttt{mathoffm@adobe.com} \\
\And
Gautham J. Mysore \\
Adobe Research \\
San Francisco, CA 94103\\
\texttt{gmysore@adobe.com} \\
}

\nipsfinalcopy 

\begin{document}

\maketitle

\begin{abstract}
We propose the product-of-filters (PoF) model, a generative model that
decomposes audio spectra as sparse linear combinations of ``filters''
in the log-spectral domain. PoF makes similar assumptions to those used
in the classic homomorphic filtering approach to signal processing,
but replaces decompositions built of basic signal
processing operations with a learned decomposition based on
statistical inference. This paper formulates the PoF model and derives
a mean-field method for posterior inference and a variational EM
algorithm to estimate the model's free parameters. We demonstrate PoF's
potential for audio processing on a bandwidth expansion task, and show
that PoF can serve as an effective unsupervised feature extractor for a
speaker identification task.
\end{abstract}

\section{Introduction}
Some of the most successful approaches to audio signal processing of
the last fifty years have been based on decomposing complicated systems
into an excitation signal and some number of simpler linear systems.
One of the simplest (and most widely used) examples is linear
predictive coding (LPC), which uses a simple autoregressive model to
decompose an audio signal into an excitation
signal and linear filter \cite{lpc}. More broadly, homomorphic filtering methods
 such as cepstral analysis \cite{oppenheim1968homomorphic}
try to decompose complicated linear systems into a set of simpler
linear systems that can then be analyzed, interpreted, and manipulated
independently.

One reason that this broad approach has been successful is because it is consistent with
the way many real-world objects actually generate sound. An important
example is the human voice: human vocal sounds are generated by
running vibrations generated by the vocal folds through the rest of
the vocal tract (tongue, lips, jaw, etc.), which approximately linearly filters the
vibrations that come from the larynx or lungs.

Traditional approaches typically rely on hand-designed decompositions
built of basic operations such as Fourier transforms, discrete cosine
transforms, and least-squares solvers. In this paper we take a more
data-driven approach, and derive a generative product-of-filters (PoF)
model that \emph{learns} a statistical-inference-based decomposition
that is tuned to be appropriate to the data being analyzed. Like
traditional homomorphic filtering approaches, PoF decomposes audio
spectra as linear combinations of filters in the log-spectral
domain. Unlike previous approaches, these filters are learned from
data rather than selected from convenient families such as orthogonal
cosines, and the PoF model learns a sparsity-inducing prior that
prefers decompositions that use relatively few filters to explain each
observed spectrum. The result when applied to speech data is that PoF
discovers some filters that model excitation signals and some that
model the various filtering operations that the vocal tract can
perform. Given a set of these learned filters, PoF can infer how much
each filter contributed to a given audio magnitude spectrum, resulting
in a sparse, compact, interpretable representation.

The rest of the paper proceeds as follows. First, we formally
introduce the product-of-filters (PoF) model, and give more rigorous
intuitions about the assumptions that it makes. Next, we review some
previous work and show how it relates to the PoF model. Then, we derive
a mean-field variational inference algorithm that allows us to do
approximate posterior inference in PoF, as well as a variational EM
algorithm that fits the model's free parameters to data. Finally, we
demonstrate PoF's potential for audio processing on a bandwidth
expansion task, where it achieves better results than a recently
proposed technique based on non-negative matrix factorization (NMF). We also
evaluate PoF as an unsupervised feature extractor, and find that the
representation learned by the model yields higher accuracy on a
speaker identification task than the widely used mel-frequency
cepstral coefficient (MFCC) representation.

\section{Product-of-Filters Model}\label{sec:sf_prior}

We are interested in modeling audio spectrograms, which are collections of Fourier magnitude 
spectra $\mathbf{W}$ taken from some set of audio signals, where
$\mathbf{W}$ is an $F\times T$ non-negative matrix; the cell $W_{ft}$
gives the magnitude of the audio signal at frequency bin $f$
and time window (often called a frame) $t$. Each column of
$\mathbf{W}$ is the magnitude of the
fast Fourier transform (FFT) of a short window of an audio
signal, within which the spectral characteristics of the signal are
assumed to be roughly stationary.

The motivation for our model comes from the widely used homomorphic
filtering approach to audio and speech signal processing
\cite{oppenheim1968homomorphic}, where a short window of audio $w[n]$
is modeled as a convolution between an excitation signal $e[n]$ (which
might come from a speaker's vocal folds) and the impulse response
$h[n]$ of a series of linear filters (such as might be implemented by
a speaker's vocal tract):
\begin{equation}
\label{eq:homomorphic}
w[n] = (e \ast h)[n]
\end{equation}
In the spectral domain after taking the FFT,
this is equivalent to:
\begin{equation}\label{eq:convolve}
|\mathcal{W}[k]| = |\mathcal{E}[k]| \circ |\mathcal{H}[k]| 
= \exp\{\log |\mathcal{E}[k]| + \log |\mathcal{H}[k]|\}
\end{equation}
where $\circ$ denotes element-wise multiplication and $|\cdot|$
denotes the magnitude of a complex value produced by the FFT. Thus,
the convolution between these two signals corresponds to a simple addition of
their log-spectra. Another attractive feature is the symmetry between
the excitation signal $e[n]$ and the impulse response $h[n]$ of the
vocal-tract filter---convolution commutes, so mathematically (if not
physiologically) the vocal tract could just as well be exciting the
``filter'' implemented by vocal folds.

We will likewise model the observed magnitude spectra as a
product of filters. We assume each observed log-spectrum is approximately obtained
by linearly combining elements from a pool of $L$ log-filters\footnote{We will use the
  term ``filter'' when referring to $\mathbf{U}$ for the rest of the paper.}
$ \mathbf{U} \equiv [\bm{u}_1| \bm{u}_2 | \cdots | \bm{u}_L]  \in \mathbb{R}^{F\times L}$:
\begin{equation}
\log W_{ft} \approx \textstyle{\sum_l} U_{fl}a_{lt},
\end{equation}
where $a_{lt}$ denotes the activation of filter $\bm{u}_l$ in frame
$t$. We will impose some sparsity on the activations, to allow us to
encode the intuition that not all filters are always active.  This
assumption expands on the expressive power of the simple
excitation-filter model of equation \ref{eq:homomorphic}; we could
recover that model by partitioning the filters into ``excitations''
and ``vocal tracts'', requiring that exactly one ``excitation filter''
be active in each frame, and combining the weighted effects of all
``vocal tract filters'' into a single filter.

We have two main reasons for relaxing the classic excitation-filter
model to include more than two filters, one computational and one
statistical. The statistical rationale is that the parameters that
define the human voice (pitch, tongue position, etc.) are inherently
continuous, and so a very large dictionary of excitations and filters
might be necessary to explain observed inter- and intra-speaker
variability with the classic model. The computational rationale is
that clustering models (which might try to determine which excitation
is active) can be more fraught with local optima than factorial models
such as ours (which tries to determine how active each filter is), and
there is some precedent for relaxing clustering models into factorial
models \cite{ding2004k}.

Formally, we define the product-of-filters model:
\begin{equation} \label{eq:model}
\begin{split}
a_{lt} &\sim \text{Gamma}(\alpha_l, \alpha_l)\\
W_{ft} &\sim \text{Gamma}\Big(\gamma_f, \gamma_f / \exp(\textstyle{\sum_l} U_{fl} a_{lt})\Big) 
\end{split}
\end{equation}
where $\gamma_f$ is the frequency-dependent noise level. We restrict
the activations $\bm{a}_t$ (but not the filters
$\bm{u}_l$) to be non-negative; if we allowed negative $a_{lt}$, then
the the filters would be inverted, reducing the interpretability of
the model.

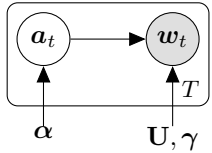
\begin{figure}[ht]
  \centering
%
%
%
%


\begin{tikzpicture}

  \node[obs] 			(w) {$\bm{w}_t$};
  \node[latent, left=of w] (a) {$\bm{a}_t$};
  \node[const, below=.8cm of w] (u) {$\mathbf{U}, \bm{\gamma}$};
  \node[const, below=.8cm of a] (alpha) {$\bm{\alpha}$};

  \edge {a} {w} ; %
  \edge {u} {w};
  \edge {alpha}{a};

  \plate {wa} {(w)(a)} {$T$};

\end{tikzpicture}
  \caption{Graphical model representation of the PoF model.}
\label{fig:sf_prior}
\end{figure}

Under this model
\begin{equation}
\begin{split}
\mathbb{E}[a_{lt}] &= 1\\
\mathbb{E}[W_{ft}] &= \exp(\textstyle{\sum_l} U_{fl} a_{lt}).
\end{split}
\end{equation}
For $l \in \{1, 2, \cdots, L\}$, $\alpha_l$ controls the sparseness of
the activations associated with filter $\bm{u}_l$; smaller values of 
$\alpha_l$ indicate that filter $\bm{u}_l$ is used more
rarely. From a generative point of view, one can view the
model as first drawing activations $a_{lt}$ from a sparse prior, then
applying multiplicative gamma noise with expected value $1$ to the
expected value $\exp(\sum_l U_{fl} a_{lt})$. A graphical model
representation of the PoF model is shown in Figure
\ref{fig:sf_prior}. 

In this paper we focus on speech applications, but the homomorphic filtering approach has also been successfully applied to model other
kinds of sounds such as musical instruments. For example,  \cite{karjalainen1993towards}  treat the effect of the random excitation, string, and body as a chain of linear systems, which can therefore be modeled as a product of filters.

\section{Related Work}
The PoF model can be interpreted as a matrix factorization model,
where we are trying to decompose the log-spectrogram. A closely
related model is non-negative matrix factorization (NMF)
\cite{seung2001algorithms} and its variations, e.g., NMF with
sparseness constrains \cite{hoyer2004non}, convex NMF
\cite{ding2010convex}, and fully Bayesian NMF
\cite{cemgil2009bayesian}. In NMF, a $F \times T$ non-negative matrix
$\mathbf{W}$ is approximately decomposed into the product of two
non-negative matrices: a $F\times K$ matrix $\mathbf{V}$ (often called
the dictionary) and a $K \times T$ matrix $\mathbf{H}$ (often called
the activations).  NMF is widely used to analyze audio
spectrograms \cite{neco09, smaragdis2003non}, largely due to its
additivity property and the parts-based representation that it
induces. It also often provides a semantically meaningful
interpretation. For example, given the NMF decomposition of a piano
sonata, the components in the dictionary are likely to correspond to
notes with different pitches, and the activations will indicate when
and how strongly each note is played. NMF's ability to isolate energy
coming from different sources in mixed recordings has made it a
popular tool for addressing source separation problems.

Although both models decompose audio spectra using a linear
combination of dictionary elements, NMF and PoF make fundamentally
different modeling assumptions. NMF models each frame of a
spectrogram as an additive combination of dictionary elements, which
approximately corresponds to modeling the corresponding time-domain
signal as a summation of parts. On the other hand, PoF models each
 frame of the spectrogram as a product of filters (sum of
log-filters), which corresponds to modeling the corresponding
time-domain signal as a convolution of filters. NMF is well suited to
decomposing \emph{polyphonic} sounds into mixtures of independent
sources, whereas PoF is well suited to decomposing \emph{monophonic}
sounds into simpler systems.

In the compressive sensing literature, there has been a great deal of
work on matrix factorization and dictionary learning by solving an
optimization problem with sparseness constraints---adding $\ell^1$
norm penalty as a convex relaxation of $\ell^0$ norm penalty
\cite{donoho2003optimally}. Online algorithms have also been proposed to
handle large data sets \cite{mairal2010online}. In principle, we could
have formulated the PoF model similarly, using an $\ell^1$ penalty and
convex optimization in place of a gamma prior and Bayesian inference.
However, in such a formulation it is unclear how we might fit the $L$
hyperparameters $\bm{\alpha}$ controlling the amount of sparsity in
the activations associated with each filter. We found that PoF best
explains speech training data when each filter $\bm{u}_l$ has its own
sparsity hyperparameter $\alpha_l$, and performing cross-validation to
select so many hyperparameters would be impractical.

\section{Inference and Parameter Estimation}
\label{sec:inference}
From Figure \ref{fig:sf_prior}, we can see that there are two computational problems that will arise when using the PoF model. First, given fixed $\mathbf{U}, \bm{\alpha}$, and $\bm{\gamma}$ and input spectrum $\bm{w}_t$, we must (approximately) compute the posterior distribution $p(\bm{a}_t | \bm{w}_t, \mathbf{U}, \bm{\alpha}, \bm{\gamma})$. This will enable us to fit the PoF model to unseen data and obtain a different representation in the latent filter space. Second, given a collection of training spectra $\mathbf{W} = \{\bm{w}_t\}^{1:T}$, we want to find the maximum likelihood estimates of the free parameters $\mathbf{U}$, $\bm{\alpha}$, and $\bm{\gamma}$. In this section, we will tackle these two problems respectively. The detailed derivations can be found in the appendix. The source code in Python is available on Github\footnote{\url{https://github.com/dawenl/pof}}.

\subsection{Mean-Field Posterior Inference}\label{sec:e-step}
The posterior $p(\bm{a}_t | \bm{w}_t, \mathbf{U}, \bm{\alpha}, \bm{\gamma})$ is intractable to compute due to the nonconjugacy of the model. Therefore, we employ mean-field variational inference \cite{jordan1999introduction}.

Variational inference is a deterministic alternative to Markov Chain Monte Carlo (MCMC) methods.  The basic idea behind variational
inference is to choose a tractable family of variational distributions
$q(\bm{a}_t)$ to approximate the intractable posterior $p(\bm{a}_t |
\bm{w}_t, \mathbf{U}, \bm{\alpha}, \bm{\gamma})$, so that the
Kullback-Leibler (KL) divergence between the variational distribution and
the true posterior $\text{KL}(q_a \| p_{a|\mathbf{W}})$ is
minimized. In particular, we are using the mean-field family which is
completely factorized, i.e., $q(\bm{a}_t) = \prod_l q(a_{lt})$. For
each $a_{lt}$, we choose a variational distribution from
the same family as $a_{lt}$'s prior distribution: $q(a_{lt}) =
\text{Gamma}(a_{lt}; \nu_{lt}^a, \rho_{lt}^a)$. $\bm{\nu}^a_t$ and
$\bm{\rho}^a_t$ are free parameters that we will tune to minimize the
KL divergence between $q$ and the posterior.

We can lower bound the marginal likelihood of the input spectrum $\bm{w}_t$: 
\begin{align} \label{eq:lbound}
&\log p(\bm{w}_t | \mathbf{U}, \bm{\alpha}, \bm{\gamma})\nonumber \\ 
\geq &~\mathbb{E}_q [\log p(\bm{w}_t, \bm{a}_t| \mathbf{U}, \bm{\alpha}, \bm{\gamma})] - \mathbb{E}_q [\log q(\bm{a}_t)] \overset{\bigtriangleup}{=} \mathcal{L}(\bm{\nu}^a_t, \bm{\rho}^a_t)
\end{align}
To compute the variational lower bound $\mathcal{L}(\bm{\nu}^a_t, \bm{\rho}^a_t)$, the necessary expectations $\mathbb{E}_q[a_{lt}] = \nu_{lt}^a / \rho_{lt}^a$ and $\mathbb{E}_q[\log a_{lt}] =  \psi(\nu_{lt}^a) - \log \rho_{lt}^a$, where $\psi(\cdot)$ is the digamma function, are both easy to compute. For $\mathbb{E}_q[\exp(-U_{fl} a_{lt})]$, we seek for the moment-generating function of a gamma-distributed random variable and obtain the expectation as:
\begin{equation}
\mathbb{E}_q[\exp(-U_{fl} a_{lt})] = \left(1 + \textstyle{\frac{U_{fl}}{\rho^a_{lt}}}\right)^{-\nu_{lt}^a}
\end{equation}
for $U_{fl} > -\rho_{lt}^a$, and $+\infty$ otherwise\footnote{Technically the expectation for $U_{fl} \leq -\rho_{lt}^a$ is undefined. Here we treat it as $+\infty$ so that when $U_{fl} \le -\rho_{lt}^a$ the variational lower bound goes to $-\infty$ and the optimization can be carried out seamlessly.}.

The nonconjugacy and the exponents in the likelihood model preclude optimizing the lower bound by traditional closed-form coordinate ascent updates. Instead, we compute the gradient of $\mathcal{L}(\bm{\nu}^a_t, \bm{\rho}^a_t)$ with respect to variational parameters $\bm{\nu}_{t}^a$ and $\bm{\rho}_{t}^a$ and use Limited-memory BFGS (L-BFGS) to optimize the variational lower bound, which guarantees to find a local optimum and optimal variational parameters $\{\hat{\bm{\nu}}_t^{a}, \hat{\bm{\rho}}_t^{a}\}$. 

Note that in the posterior inference, the optimization problem is independent for different frame $t$. Therefore, given input spectra $\{\bm{w}_t\}^{1:T}$, we can break down the whole problem into $T$ independent sub-problems which can be solved in parallel. 

\subsection{Parameter Estimation}

Given a collection of training audio spectra $\{\bm{w}_t\}^{1:T}$, we carry out parameter estimation for the PoF model by finding the maximum likelihood estimates of the free parameters $\mathbf{U}, \bm{\alpha}$, and $\bm{\gamma}$, approximately marginalizing out $\bm{a}_t$.

We formally define the parameter estimation problem as
\begin{align} \label{eq:obj}
\hat{\mathbf{U}}, \hat{\bm{\alpha}}, \hat{\bm{\gamma}} &= \argmax_{\mathbf{U}, \bm{\alpha}, \bm{\gamma}} \sum_t \log p(\bm{w}_t | \mathbf{U}, \bm{\alpha}, \bm{\gamma}) \\ 
&= \argmax_{\mathbf{U}, \bm{\alpha}, \bm{\gamma}} \sum_t \log \int_{\bm{a}_t} p(\bm{w}_t, \bm{a}_t | \mathbf{U}, \bm{\alpha}, \bm{\gamma}) \mathrm{d} \bm{a}_t \nonumber
\end{align}
This problem can be solved by variational Expectation-Maximization (EM) which maximizes a lower bound on marginal likelihood in equation \ref{eq:lbound} with respect to the variational parameters, and then for the fixed values of variational parameters, maximizes the lower bound with respect to the model's free parameters $\mathbf{U}$, $\bm{\alpha}$, and $\bm{\gamma}$. 

\paragraph{E-step} For each $\bm{w}_t$ where $t = 1, 2, \cdots, T$, perform posterior inference by optimizing the values of the variational parameters $\{\hat{\bm{\nu}}_t^{a}, \hat{\bm{\rho}}_t^{a}\}$. This is done as described in Section \ref{sec:e-step}.

\paragraph{M-step}
Maximize the variational lower bound in equation \ref{eq:lbound}, which is equivalent to maximizing the following objective: 
\begin{equation} \label{eq:obj_m}
\mathcal{Q}(\mathbf{U}, \bm{\alpha}, \bm{\gamma}) = \textstyle{\sum_t} \mathbb{E}_q [ \log p(\bm{w}_t, \bm{a}_t | \mathbf{U}, \bm{\alpha}, \bm{\gamma}) ] 
\end{equation}
This is accomplished by finding the maximum likelihood estimates using
the expected sufficient statistics for each $\bm{a}_t$ that were
computed in the E-step. There are no simple closed-form updates for the
M-step. Therefore, we compute the gradient of $\mathcal{Q}(\mathbf{U},
\bm{\alpha}, \bm{\gamma})$ with respect to $\mathbf{U}, \bm{\alpha},
\bm{\gamma}$, respectively, and use L-BFGS to optimize the bound in
equation \ref{eq:obj_m}.

The most time-consuming part for M-step is updating $\mathbf{U}$,
which is a $F\times L$ matrix. However, note that the optimization
problem is independent for different frequency bins $f\in \{1, 2,
\cdots, F\}$. Therefore, we can update $\mathbf{U}$ by optimizing each
row independently, and in parallel if desired.

\section{Evaluation}

\begin{figure}
        \centering
        \begin{subfigure}[b]{1\textwidth}
                \includegraphics[width=\textwidth]{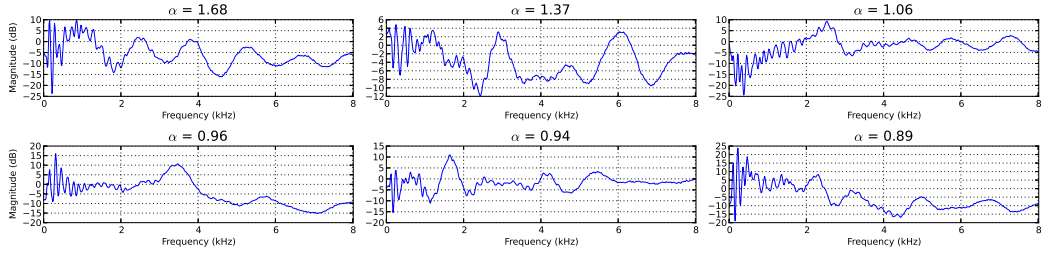}		
		\caption{The top 6 filters $\bm{u}_l$ with the largest $\alpha_l$ values (shown above each plot). }
		\label{fig:filters}
        \end{subfigure}%
       \\
        \begin{subfigure}[b]{1\textwidth}
                \includegraphics[width=\textwidth]{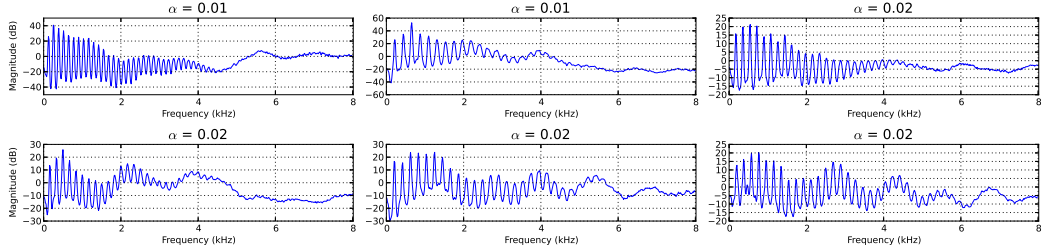}
		\caption{The top 6 filters $\bm{u}_l$ with the smallest $\alpha_l$ values (shown above each plot). }
		\label{fig:sources}
        \end{subfigure}
        \caption{The representative filters learned from the PoF model with $L = 50$.}\label{fig:sf}
\end{figure}

We conducted experiments to assess the PoF model on two different tasks. We evaluate PoF's ability to infer missing data in the bandwidth expansion task. We also explore the potential of the PoF model as an unsupervised feature extractor for the speaker identification task. 

Both tasks require pre-trained parameters $\mathbf{U}$, $\bm{\alpha}$, and $\bm{\gamma}$, which we learned from the TIMIT Speech Corpus \cite{fisher1986darpa}. It contains speech sampled at 16000 Hz from 630 speakers of eight major dialects of American English, each reading ten phonetically rich sentences. The parameters were learned from 20 randomly selected speakers (10 males and 10 females). We performed a $1024$-point FFT with Hann window and $50\%$ overlap, thus the number of frequency bins was $F = 513$. We performed the experiments on magnitude spectrograms except where specified otherwise. 

We tried different model orders $L \in \{10, 20, \cdots, 80\}$ and
evaluated the lower bound on the marginal likelihood $\log p(\bm{w}_t
| \mathbf{U}, \bm{\alpha}, \bm{\gamma})$ in equation
\ref{eq:lbound}. In general, larger $L$ will give us a larger
variational lower bound and will be slower to train. 
In our experiments, we set $L = 50$ as a compromise between performance 
and computational efficiency. 
We initialized all the variational parameters $\bm{\nu}_t^a$ and $\bm{\rho}_t^a$ with random draws from a gamma distribution with shape parameter 100 and inverse-scale parameter 100. This yields a diffuse and smooth initial variational posterior, which helped avoid bad local optima.
We ran variational EM until the variational lower bound increased by less than 0.01\%. 

Figure \ref{fig:sf} demonstrates some representative filters learned from the PoF model with $L = 50$.
The six filters $\bm{u}_l$ associated with the largest values of
$\alpha_l$ are shown in Figure \ref{fig:filters}, and the six filters
associated with the smallest values of $\alpha_l$ are shown in Figure
\ref{fig:sources}. Small values of $\alpha_l$ indicate a prior
preference to use the associated filters less frequently, since the
$\mathrm{Gamma}(\alpha_l,\alpha_l)$ prior places more mass near 0 when
$\alpha_l$ is smaller. The filters in Figure \ref{fig:sources}, which
are used relatively rarely, tend to have the strong harmonic structure
displayed by the log-spectra of periodic signals, while the filters in
Figure \ref{fig:filters} tend to vary more smoothly, suggesting that
they are being used to model the filtering induced by the vocal tract.
The periodic ``excitation'' filters tend to be used more rarely, which is consistent with the fact that normally there is not more than one excitation signal contributing to a speaker's voice. (Very few people can speak or sing more than one pitch simultaneously.) The model has more freedom to use several of the coarser ``vocal tract'' filters per spectrum, which is consistent with the fact that several aspects of the vocal tract may be combined to filter the excitation signal generated by a speaker's vocal folds.

Despite the non-convexity inherent to all dictionary-learning problems, which causes the details of the filters vary from run to run, training with multiple random restarts and different speakers produced little impact on the filters that the PoF model learned; in all cases with different model order $L$, we found the same ``excitation/filter'' structure, similar to what is shown in Figure \ref{fig:sf}.

\subsection{Bandwidth Expansion}

We demonstrate PoF's potential in audio processing applications on a
bandwidth expansion task, where the goal is to infer the contents of a
full-bandwidth signal given only the contents of a band-limited
version of that signal. Bandwidth expansion has applications to
restoration of low-quality audio such as might be recorded from a
telephone or cheap microphone.

Given the parameters $\mathbf{U}, \bm{\alpha},$ and $\bm{\gamma}$ fit
to full-bandwidth training data, we can treat the bandwidth expansion
problem as a missing data problem. Given spectra from a band-limited
recording $\mathbf{W^{\text{bl}}} = \{\bm{w}^{\text{bl}}_t\}^{1:T}$,
the model implies a posterior distribution
$p(\bm{a}_t|\bm{w^{\text{bl}}_t})$ over the activations $\bm{a}_t$
associated with the band-limited signal, for $t = \{1, 2, \cdots, T\}$. 
We can approximate this posterior
using the variational inference algorithm from Section \ref{sec:e-step} by only using the band-limited part of
$\mathbf{U}$ and $\bm{\gamma}$. Then we can reconstruct the full-bandwidth spectra
by combining the inferred $\{\bm{a}_t\}^{1:T}$ with the full-bandwidth
$\mathbf{U}$. Following the model formulation in equation
\ref{eq:model}, we might either estimate the full-bandwidth spectra
using
\begin{equation}
\mathbb{E}_q [W^{\text{fb}}_{ft}] = \textstyle{\prod_l} \mathbb{E}_q [\exp(U_{fl} a_{lt})]
\end{equation}
or
\begin{equation}
\label{eq:bwreconstruction}
\mathbb{E}_q [W^{\text{fb}}_{ft}] = \exp \{\textstyle{\sum_l} U_{fl} \cdot \mathbb{E}_q [a_{lt}]\}.
\end{equation}
We use equation \ref{eq:bwreconstruction}, both because it is more
stable and because human auditory perception is logarithmic; if we are
summarizing the posterior distribution with a point estimate, the
expectation on the log-spectral domain is more perceptually natural.

As a comparison, NMF is widely used for bandwidth expansion \cite{bansal2005bandwidth, raj2007bandwidth, sun2013non}. The full-bandwidth training spectra $\mathbf{W}^{\text{train}}$, which are also used to learn the parameters $\mathbf{U}$, $\bm{\alpha}$, and $\bm{\gamma}$ for the PoF model, are decomposed by NMF as $\mathbf{W}^{\text{train}} \approx \mathbf{V} \mathbf{H}$, where $\mathbf{V}$ is the dictionary and $\mathbf{H}$ is the activation. Then given the band-limited spectra $\mathbf{W^{\text{bl}}}$, we can use the band-limited part of $\mathbf{V}$ to infer the activation $\mathbf{H}^{\text{bl}}$. Finally, we can reconstruct the full-bandwidth spectra by computing $\mathbf{V}\mathbf{H}^{\text{bl}}$. 

Based on how the loss function is defined, there can be different types of NMF models: KL-NMF \cite{seung2001algorithms} which is based on Kullback-Leibler divergence, and IS-NMF \cite{neco09} which is based on Itakura-Saito divergence, are among the most commonly used NMF decomposition models in audio signal processing. We compare the PoF model with both KL-NMF and IS-NMF with different model orders $K$ = 25, 50, and 100. We used the standard multiplicative updates for NMF and stopped the iterations when the decrease in the cost function was less than $0.01\%$. For IS-NMF, we used power spectra instead of magnitude spectra, since the power spectrum representation is more consistent with the statistical assumptions that underlie the Itakura-Saito divergence. 

We randomly selected 10 speakers (5 male and 5 female) from TIMIT that do not overlap with the speakers used to fit the model parameters $\mathbf{U}$, $\bm{\alpha}$, and $\bm{\gamma}$, and took 3 sentences from each speaker as test data. We cut off all the contents below 400 Hz and above 3400 Hz to obtain band-limited recordings of approximately telephone-quality speech. 

In previous NMF-based bandwidth expansion work \cite{bansal2005bandwidth, raj2007bandwidth, sun2013non}, all experiments are done in a speaker-dependent setting, which means the model is trained from the target speaker. What we are doing here, on the other hand, is speaker-independent: we use no prior knowledge about the specific speaker whose speech is being restored\footnote{When we conducted speaker-dependent
experiments, both PoF and NMF produced nearly ceiling-level results. Thus we only report results on the harder and more practically relevant speaker-independent problem.}. To our knowledge, little if any work has been done on speaker-independent bandwidth expansion based on NMF decompositions. 

To evaluate the quality of the reconstructed recordings, we used the
composite objective measure \cite{hu2008evaluation} and short-time
objective intelligibility \cite{taal2011algorithm} metrics. These
metrics measure different aspects of the ``distance'' between the
reconstructed speech and the original speech. The composite objective
measure (will be abbreviated as OVRL, as it reflects the overall sound
quality) was originally proposed as a quality measure for speech
enhancement. It aggregates different basic objective measures and has
been shown to correlate with humans' perceptions of audio
quality. OVRL is based on the predicted perceptual auditory rating and
is in the range of 1 to 5 (1: bad; 2: poor; 3: fair; 4: good; 5:
excellent). The short-time objective intelligibility measure (STOI) is
a function of the clean speech and reconstructed speech, which
correlates with the intelligibility of the reconstructed speech, that
is, it predicts the ability of listeners to understand what words are
being spoken rather than perceived sound quality. STOI is computed as
the average correlation coefficient from 15 one-third octave bands
across frames, thus theoretically should be in the range of -1 to 1,
where larger values indicate higher expected intelligibility. However,
in practice, even when we filled out the missing contents with random
noise, the STOI is 0.306 $\pm$ 0.016, which can be interpreted as a practical
lower bound on the test data.

The average OVRL and STOI with two standard errors\footnote{For both
  OVRL and STOI, we used the MATLAB implementation from the original
  authors.}  across 30 sentences for different methods, along with
these from the band-limited input speech as baseline, are reported in
Table \ref{tab:bwe}. We can see that NMF improves STOI by a small amount, and PoF
improves it slightly more, but the improvement in both cases is fairly
small. This may be because the band-limited input speech already has a
relatively high STOI (telephone-quality speech is fairly
intelligible).  On the other hand, PoF produces better
predicted perceived sound quality as measured by OVRL than KL-NMF and
IS-NMF by a large margin regardless of the model order $K$, improving
the sound quality from poor-to-fair (2.71 OVRL) to fair-to-good (3.25 OVRL).

\begin{table} 
\caption{ Average OVRL (composite objective measure) and STOI (short-time objective intelligibility) score with two standard errors (in parenthesis) for the bandwidth expansion task from different methods. OVRL is in the range of 1 to 5 [1: bad; 2: poor; 3: fair; 4: good; 5: excellent]. STOI is the average correlation coefficient, thus theoretically should be in the range of -1 to 1, where larger values indicate higher expected intelligibility. }
\begin{center}
\begin{tabular}{   c  c |  c | c |}
\cline{3-4}
& & OVRL & STOI  \\
\hline
\multicolumn{2}{|c|}{Band-limited input} & 1.72 (0.16) & 0.762 (0.012)\\
\hline
 \multicolumn{1}{ |c| }{\multirow{3}{*}{KL-NMF} }& $K$=25 & 2.60 (0.12) & 0.786 (0.013)  \\
 \cline{2-4}
 \multicolumn{1}{|c|}{} & $K$=50 & 2.71 (0.14)  & 0.790 (0.013)\\
  \cline{2-4}
  \multicolumn{1}{|c|}{}  & $K$=100 & 2.41 (0.10)  & 0.759 (0.012) \\
\hline 
\multicolumn{1}{ |c| }{\multirow{3}{*}{IS-NMF} }& $K$=25 & 2.43 (0.15)  & 0.779 (0.013)\\
 \cline{2-4}
  \multicolumn{1}{|c|}{}  & $K$=50 & 2.62 (0.12)  & 0.774 (0.014) \\
  \cline{2-4}
  \multicolumn{1}{|c|}{}  & $K$=100 & 2.15 (0.10)  & 0.751 (0.012)\\
  \hline
  \multicolumn{2}{|c|}{PoF} & \textbf{3.25 (0.13)} & \textbf{0.804 (0.010)}
  \\ 
\hline
\end{tabular}\label{tab:bwe}
\end{center}
\end{table}

\subsection{Feature Learning and Speaker Identification}
We explore PoF's potential as an unsupervised feature extractor. One way to interpret the PoF model is that it attempts to represent the data in a latent filter space. Therefore, given spectra $\{\bm{w}_t\}^{1:T}$, we can use the coordinates in the latent filter space $\{\bm{a}_t\}^{1:T}$ as features. Since we believe the inter- and intra-speaker variability is well-captured by the PoF model, we use speaker identification to evaluate the effectiveness of these features.

We compare our learned representation with mel-frequency cepstral
coefficients (MFCCs), which are widely used in various speech and
audio processing tasks including speaker identification. MFCCs are
computed by taking the discrete cosine transform (DCT) on mel-scale
log-spectra and using only the low-order coefficients. PoF can be
understood in similar terms; we are also trying to explain the
variability in log-spectra in terms of a linear combination of
dictionary elements. However, instead of using the fixed, orthogonal
DCT basis, PoF learns a filter space that is tuned to the statistics
of the input. Therefore, it seems reasonable to hope that the
coefficients $\bm{a}_t$ from the PoF model, which will be abbreviated
as PoFC, might compare favorably with MFCCs as a feature
representation.

We evaluated speaker identification under the following scenario: identify different speakers from recordings where each speaker may start and finish talking at random time, but at any given time there is only one speaker speaking (like during a meeting). This is very
similar to speaker diarization \cite{Fox:AOAS2011}, but here we assume
we know \emph{a priori} the number of speakers and certain amount of training data is available for each speaker.  Our goal in this experiment was to demonstrate that
the PoFC representation captures information that is missing or difficult to extract from the MFCC
representation, rather than trying to build a state-of-the-art speaker identification system.

We randomly selected 10 speakers (5 males and 5 females) from TIMIT outside the training data we used to learn the free parameters $\mathbf{U}$, $\bm{\alpha}$, and $\bm{\gamma}$. We used the first 13 DCT coefficients which is a standard choice for computing MFCC. We obtained the PoFC by doing posterior inference as described in Section \ref{sec:e-step} and used $\mathbb{E}_q[\bm{a}_t]$ as a point estimate summary. For both MFCC and PoFC, we computed the first-order and second-order differences and concatenated them with the original feature. 

We treat speaker identification problem as a classification problem
and make predictions for each frame. We trained a multi-class
(one-vs-the-rest) linear SVM using eight sentences from each speaker
and tested with the remaining two sentences, which gave us about 7800
frames of training data and 1700 frames of test data. The test
data was randomly permuted so that the order in which sentences appear
is random, which simulates the aforementioned scenario.

The frame level accuracy is reported in the first row of Table
\ref{tab:spk}. We can see PoFC increases the accuracy by a large
margin (from $49.1\%$ to $60.5\%$). To make use of temporal
information, we used a simple median filter smoother with length 25,
which boosts the performance for both representations equally; these
results are reported in the second row of Table \ref{tab:spk}.

Although MFCCs and PoFCs capture similar information, concatenating
both sets of features yields better accuracy than that obtained by
either feature set alone. The results achieved by combining the
features are summarized in the last column of Table \ref{tab:spk},
which indicates that MFCCs and PoFCs capture complementary
information. These results, which use a relatively simple frame-level
classifier, suggest that PoFC could produce even better accuracy when
used in a more sophisticated model (e.g. a deep neural network).

\begin{table}[t]
\caption{10-speaker identification accuracy using PoFC, MFCC, and combined. The first row shows the raw frame-level test accuracy. The second row shows the result after applying a simple median filter with length 25 on the frame-level prediction.}
\begin{center}
\begin{tabular}{| c || c | c | c |}
  \hline                        
   & MFCC & PoFC & MFCC + PoFC \\ \hline
  Frame-level & $49.1\%$ & $60.5\%$  & $65.5\%$\\ \hline
  Smoothing  & $74.2\%$ & $85.0\%$ & $89.5\%$\\
  \hline  
\end{tabular}\label{tab:spk}
\end{center}
\end{table}

\section{Discussion and Future Work}
In this paper, we proposed the product-of-filters (PoF) model, a
generative model which makes similar assumptions to those used in the
classic homomorphic filtering approach to signal processing. We
derived variational inference and parameter estimation algorithms,
and demonstrated experimental improvements on a
bandwidth expansion task and showed that PoF can serve as an effective
unsupervised feature extractor for speaker identification.

In this paper, we derived PoF as a standalone model. However, it can also be used as
a building block and integrated into a larger model, e.g., as a prior
for the dictionary in a probabilistic NMF model.

Although the optimization in the variational EM algorithm can be parallelized,
currently we cannot fit PoF to large-scale speech data on a single
machine. Leveraging recent developments in stochastic variational
inference \cite{hoffman2013stochastic}, it would be possible to learn
the free parameters from a much larger, more diverse speech corpus, or
even from streams of data.

\bibliography{pof}

\begin{thebibliography}{10}

\bibitem{bansal2005bandwidth}
D.~Bansal, B.~Raj, and P.~Smaragdis.
\newblock Bandwidth expansion of narrowband speech using non-negative matrix
  factorization.
\newblock In {\em INTERSPEECH}, pages 1505--1508, 2005.

\bibitem{cemgil2009bayesian}
A.~T. Cemgil.
\newblock Bayesian inference for nonnegative matrix factorisation models.
\newblock {\em Computational Intelligence and Neuroscience}, 2009.

\bibitem{ding2004k}
C.~Ding and X.~He.
\newblock {$K$}-means clustering via principal component analysis.
\newblock In {\em Proceedings of the International Conference on Machine
  learning}, page~29. ACM, 2004.

\bibitem{ding2010convex}
C.~Ding, T.~Li, and M.~I. Jordan.
\newblock Convex and semi-nonnegative matrix factorizations.
\newblock {\em Pattern Analysis and Machine Intelligence, IEEE Transactions
  on}, 32(1):45--55, 2010.

\bibitem{donoho2003optimally}
D.~L. Donoho and M.~Elad.
\newblock Optimally sparse representation in general (nonorthogonal)
  dictionaries via $\ell^1$ minimization.
\newblock {\em Proceedings of the National Academy of Sciences},
  100(5):2197--2202, 2003.

\bibitem{neco09}
C.~F\'{e}votte, N.~Bertin, and J.-L. Durrieu.
\newblock Nonnegative matrix factorization with the {I}takura-{S}aito
  divergence: with application to music analysis.
\newblock {\em Neural Computation}, 21(3):793--830, Mar. 2009.

\bibitem{fisher1986darpa}
W.~M. Fisher, G.~R. Doddington, and K.~M. Goudie-Marshall.
\newblock The {DARPA} speech recognition research database: specifications and
  status.
\newblock In {\em Proc. DARPA Workshop on speech recognition}, pages 93--99,
  1986.

\bibitem{Fox:AOAS2011}
E.B. Fox, E.B. Sudderth, M.I. Jordan, and A.S. Willsky.
\newblock {A Sticky HDP-HMM with Application to Speaker Diarization}.
\newblock {\em Annals of Applied Statistics}, 5(2A):1020--1056, 2011.

\bibitem{hoffman2013stochastic}
M.~Hoffman, D.~Blei, C.~Wang, and J.~Paisley.
\newblock Stochastic variational inference.
\newblock {\em Journal of Machine Learning Research}, 14:1303--1347, 2013.

\bibitem{hoyer2004non}
P.~O. Hoyer.
\newblock Non-negative matrix factorization with sparseness constraints.
\newblock {\em The Journal of Machine Learning Research}, 5:1457--1469, 2004.

\bibitem{hu2008evaluation}
Y.~Hu and P.~C. Loizou.
\newblock Evaluation of objective quality measures for speech enhancement.
\newblock {\em Audio, Speech, and Language Processing, IEEE Transactions on},
  16(1):229--238, 2008.

\bibitem{jordan1999introduction}
M.~I. Jordan, Z.~Ghahramani, T.~S. Jaakkola, and L.~K. Saul.
\newblock An introduction to variational methods for graphical models.
\newblock {\em Machine learning}, 37(2):183--233, 1999.

\bibitem{karjalainen1993towards}
M.~Karjalainen, V.~V{\"a}lim{\"a}ki, and Z.~J{\'a}nosy.
\newblock Towards high-quality sound synthesis of the guitar and string
  instruments.
\newblock In {\em Proceedings of the International Computer Music Conference},
  pages 56--56, 1993.

\bibitem{seung2001algorithms}
D.D. Lee and H.S. Seung.
\newblock Algorithms for non-negative matrix factorization.
\newblock {\em Advances in Neural Information Processing Systems}, 13:556--562,
  2001.

\bibitem{mairal2010online}
J.~Mairal, F.~Bach, J.~Ponce, and G.~Sapiro.
\newblock Online learning for matrix factorization and sparse coding.
\newblock {\em The Journal of Machine Learning Research}, 11:19--60, 2010.

\bibitem{oppenheim1968homomorphic}
A.~Oppenheim and R.~Schafer.
\newblock Homomorphic analysis of speech.
\newblock {\em Audio and Electroacoustics, IEEE Transactions on},
  16(2):221--226, 1968.

\bibitem{lpc}
D.~O'Shaughnessy.
\newblock Linear predictive coding.
\newblock {\em Potentials, IEEE}, 7(1):29--32, 1988.

\bibitem{raj2007bandwidth}
B.~Raj, R.~Singh, M.~Shashanka, and P.~Smaragdis.
\newblock Bandwidth expansionwith a {P}{\'o}lya urn model.
\newblock In {\em Acoustics, Speech and Signal Processing, IEEE International
  Conference on}, volume~4, pages IV--597. IEEE, 2007.

\bibitem{smaragdis2003non}
P.~Smaragdis and J.~C. Brown.
\newblock Non-negative matrix factorization for polyphonic music transcription.
\newblock In {\em Applications of Signal Processing to Audio and Acoustics,
  IEEE Workshop on.}, pages 177--180. IEEE, 2003.

\bibitem{sun2013non}
D.~L. Sun and R.~Mazumder.
\newblock Non-negative matrix completion for bandwidth extension: A convex
  optimization approach.
\newblock In {\em Machine Learning for Signal Processing (MLSP), IEEE
  International Workshop on}, pages 1--6. IEEE, 2013.

\bibitem{taal2011algorithm}
C.~H. Taal, R.~C. Hendriks, R.~Heusdens, and J.~Jensen.
\newblock An algorithm for intelligibility prediction of time--frequency
  weighted noisy speech.
\newblock {\em Audio, Speech, and Language Processing, IEEE Transactions on},
  19(7):2125--2136, 2011.

\end{thebibliography}
\bibliographystyle{plain}


\newpage
\appendix

\section{Variational EM for Product-of-Filters Model} \label{app:sf}

\subsection{E-step (Posterior Inference)} \label{app:e}
Following Section \ref{sec:e-step}, the variational lower bound for the E-step (equation \ref{eq:lbound}):
\begin{equation}
\begin{split}
& \log p(\bm{w}_t | \mathbf{U}, \bm{\alpha}, \bm{\gamma})\\
=&  \log \int_{\bm{a}_t} q(\bm{a}_t) \frac{p(\bm{w}_t, \bm{a}_t | \mathbf{U}, \bm{\alpha}, \bm{\gamma})}{q(\bm{a}_t)} \mathrm{d} \bm{a}_t\\
\geq&  \int_{\bm{a}_t} q(\bm{a}_t)  \log \frac{p(\bm{w}_t, \bm{a}_t | \mathbf{U}, \bm{\alpha}, \bm{\gamma})}{q(\bm{a}_t)} \mathrm{d}\bm{a}_t\\
=&~  \mathbb{E}_q [\log p(\bm{w}_t, \bm{a}_t | \mathbf{U}, \bm{\alpha}, \bm{\gamma})] - \mathbb{E}_q [\log q(\bm{a}_t)]\\
\equiv&~  \mathcal{L}(\bm{\nu}^a_t, \bm{\rho}^a_t)
\end{split}
\end{equation}
The second term is the entropy of a gamma-distributed random variable:
\begin{align*}
&-\mathbb{E}_q [\log q(\bm{a}_t)] \\
= &\sum_l \Big( \nu_{lt}^a - \log \rho_{lt}^a + \log \Gamma(\nu^a_{lt}) + (1 - \nu_{lt}^a)\psi(\nu_{lt}^a)\Big)
\end{align*}

For the first term, we can keep the parts which depend on $\{\bm{\nu}^a_t, \bm{\rho}^a_t\}$:
\begin{align*}
& \mathbb{E}_q [\log p(\bm{w}_t, \bm{a}_t | \mathbf{U}, \bm{\alpha}, \bm{\gamma})]  \\
 = &~ \mathbb{E}_q [\log p(\bm{w}_t | \bm{a}_t, \mathbf{U}, \bm{\gamma})] + \mathbb{E}_q [\log p(\bm{a}_t | \bm{\alpha})]  \\
 = & ~\text{const} + \sum_l  \Big\{ (\alpha_l - 1) \mathbb{E}_q [ \log a_{lt} ] - \alpha_l \mathbb{E}_q [ a_{lt} ] \Big\} -  \sum_f \gamma_f \Big\{W_{ft} \prod_{l} \mathbb{E}_q [ \exp(- U_{fl} a_{lt})] +\sum_l U_{fl} \mathbb{E}_q [ a_{lt}] \Big\} 
\end{align*}

Take the derivative of $\mathcal{L}(\bm{\nu}^a_t, \bm{\rho}^a_t)$ with respect to $\nu_{lt}^a$ and $\rho_{lt}^a$:
\begin{align}
\frac{\partial \mathcal{L}}{\partial \nu_{lt}^a} &= \sum_f \gamma_f \biggl\{ W_{ft}  \log(1 + \frac{U_{fl}}{\rho^a_{lt}}) \prod_{j=1}^L \mathbb{E}_q [\exp(-U_{fj} a_{jt} )] \nonumber -\frac{U_{fl}}{\rho_{lt}^a} \biggl\} + (\alpha_l - \nu_{lt}^a) \psi_1(\nu_{lt}^a) + 1 - \frac{\alpha_l}{\rho^a_{lt}}
\end{align}
\begin{align}
\frac{\partial \mathcal{L}}{\partial \rho_{lt}^a} &= \frac{\nu_{lt}}{(\rho_{lt}^a)^2} \sum_f \gamma_f \biggl\{U_{fl} -W_{ft} (1 + \frac{U_{fl}}{\rho_{lt}^a})^{-1} U_{fl} \nonumber \prod_{j=1}^L \mathbb{E}_q [\exp(- U_{fj} a_{jt} )] \biggl\} + \alpha_l (\frac{\nu_{lt}}{(\rho_{lt}^a)^2} - \frac{1}{\rho_{lt}^a})
\end{align}
where $\psi_1(\cdot)$ is the trigamma function. 

\subsection{M-step} \label{app:m}
The objective function for M-step is:
\begin{equation}
\begin{split}
\mathcal{Q}(\mathbf{U}, \bm{\alpha}, \bm{\gamma}) &= \sum_t \mathbb{E}_q [ \log p(\bm{w}_t, \bm{a}_t | \mathbf{U}, \bm{\alpha}, \bm{\gamma}) ] \\
&= \sum_t \mathbb{E}_q [\log p(\bm{w}_t | \bm{a}_t, \mathbf{U}, \bm{\gamma})] + \mathbb{E}_q [\log p(\bm{a}_t | \bm{\alpha})] 
\end{split}
\end{equation}
where
\begin{align*}
&\mathbb{E}_q [\log p(\bm{w}_t | \bm{a}_t, \mathbf{U}, \bm{\gamma})] \\
=& \sum_f \Big( \gamma_f \log \gamma_f - \gamma_ f \sum_l U_{fl} \mathbb{E}_q [ a_{lt} ]  - \log\Gamma(\gamma_f) + (\gamma_f - 1)\log W_{ft} - W_{ft} \gamma_f \prod_l \mathbb{E}_q [ \exp(- U_{fl} a_{lt} ) ] \Big)\\
& \mathbb{E}_q [\log p(\bm{a}_t | \bm{\alpha})] \\
=&\sum_l  \Big( \alpha_l \log \alpha_l - \log \Gamma(\alpha_l) + (\alpha_l - 1)\mathbb{E}_q [ \log a_{lt} ] - \alpha_l \mathbb{E}_q [ a_{lt} ] \Big)
\end{align*}
Take the derivative with respect to $\mathbf{U}, \bm{\alpha}, \bm{\gamma}$, we obtain the following gradients:
\begin{align}
\frac{\partial \mathcal{Q}}{\partial U_{fl}} &= \sum_t \Big( - \mathbb{E}_q [ a_{lt} ] + W_{ft} \mathbb{E}_q [ a_{lt}]  (1 + \frac{U_{fl}}{\rho^a_{lt}})^{-(\nu_{lt}^a+1)} \prod_{j \neq l} \mathbb{E}_q [ \exp(- U_{fj} a_{jt}) ] \Big)\\
\frac{\partial \mathcal{Q}}{\partial \alpha_l} &=  \sum_t \Big( \log \alpha_l + 1 - \psi(\alpha_l) + \mathbb{E}_q [ \log a_{lt} ] - \mathbb{E}_q [ a_{lt} ] \Big)\\
\frac{\partial \mathcal{Q}}{\partial \gamma_f} &= \sum_t \Big( \log \gamma_f - \sum_l U_{fl} \mathbb{E}_q [ a_{lt} ] + 1 - \psi(\gamma_f)  + \log W_{ft} - W_{ft} \prod_l \mathbb{E}_q [ \exp(- U_{fl} a_{lt} ) ] \Big)
\end{align}
Note that the optimization problem for $\mathbf{U}$ is independent for different frequency bin $f\in \{1, 2, \cdots, F\}$, as reflected by the gradient. 

\end{document}